\documentclass[a4paper,conference]{IEEEtran}
\IEEEoverridecommandlockouts
\usepackage{cite}
\usepackage{amsmath,amssymb,amsfonts,bm,commath}
\usepackage{algorithmic}
\usepackage{graphicx}
\usepackage{textcomp}
\usepackage{xcolor}
\usepackage[
  separate-uncertainty = true,
  multi-part-units = repeat
]{siunitx}
\usepackage[linesnumbered,ruled,noend]{algorithm2e}
\usepackage{float}

\def\BibTeX{{\rm B\kern-.05em{\sc i\kern-.025em b}\kern-.08em
    T\kern-.1667em\lower.7ex\hbox{E}\kern-.125emX}}

\renewcommand{\baselinestretch}{.98}

\begin{document}

\title{GraphHD: Efficient graph classification using hyperdimensional computing\vspace{-5pt}}

\author{

\IEEEauthorblockN{Igor Nunes, Mike Heddes, Tony Givargis, Alexandru Nicolau and Alex Veidenbaum}
\IEEEauthorblockA{\textit{Department of Computer Science} \\
UC Irvine, Irvine, USA \\
\{igord, mheddes, givargis, nicolau, alexv\}@uci.edu\vspace{-12pt}}
}

\maketitle

\begin{abstract}
Hyperdimensional Computing (HDC) developed by Kanerva is a computational model for machine learning inspired by neuroscience. HDC exploits characteristics of biological neural systems such as high-dimensionality, randomness and a holographic representation of information to achieve a good balance between accuracy, efficiency and robustness. HDC models have already been proven to be useful in different learning applications, especially in resource-limited settings such as the increasingly popular Internet of Things (IoT). One class of learning tasks that is missing from the current body of work on HDC is graph classification. Graphs are among the most important forms of information representation, yet, to this day, HDC algorithms have not been applied to the graph learning problem in a general sense. Moreover, graph learning in IoT and sensor networks, with limited compute capabilities, introduce challenges to the overall design methodology. In this paper, we present GraphHD — a baseline approach for graph classification with HDC. We evaluate GraphHD on real-world graph classification problems. Our results show that when compared to the state-of-the-art Graph Neural Networks (GNNs) the proposed model achieves comparable accuracy, while training and inference times are on average 14.6$\times$ and 2.0$\times$ faster, respectively.
\end{abstract}
\begin{IEEEkeywords}
Hyperdimensional computing, Graph classification, PageRank centrality, Graph neural networks, Graph kernels \vspace*{-5mm}
\end{IEEEkeywords}
\section{Introduction}
\label{sec:intro}



Machine Learning has played an increasingly central role in academic research and industrial applications. This popularity is due in large part to the good empirical results obtained on problems in which data is captured in the Euclidean space, such as vectors of feature values, time series data or images. However, in countless real-world scenarios, in both natural and social sciences, we are often interested in representing relationships between entities. Examples range from chemical molecules~\cite{wale2006comparison} and bioinformatics~\cite{borgwardt2005protein}, to computer vision~\cite{neumann2013graph} and analysis of social networks~\cite{yanardag2015deep}. The information about such entities and the relationships between them is inherently non-Euclidean. Graphs, instead, provide a much more natural abstraction. For this reason, the challenge of developing methodologies capable of utilizing the full potential of machine learning algorithms to deal with graphs has received a lot of attention from the scientific community in recent years.

One of the first successful strategies for graph learning problems is to calculate a measure of similarity between graphs. These methods are called graph kernels~\cite{kriege2020survey}. The similarity measurement functions are used in conjunction with kernel machines (e.g. support vector machines) to perform cognitive tasks such as classification. A myriad of graph kernel methods have been proposed, especially in the last 15 years, which will be covered in Section~\ref{sec:related}. While it is true that kernel methods are highly competitive graph learning approaches, especially on small graphs, considerable recent effort has focused on alternative  methods~\cite{morris2020tudataset} with better scaling and performance characteristics. In particular, kernel methods scale quadratically with respect to the size of the dataset and do not allow for online learning~\cite{keerthi2006building}, limiting their applicability in real-time scenarios~\cite{zhan2004increasing}. Our work introduces a new alternative approach to graph kernels.

Another popular alternative, motivated by the notorious accomplishments of deep neural networks, are graph neural networks (GNNs). GNNs are models that extend regular neural network operations, such as pooling and convolution, to handle graphs. Despite the functional accuracy achieved by GNNs, the high computational and energy cost of deep learning approaches make them difficult, or prohibitive, to be applied in real-world situations, such as those encountered in IoT and embedded applications~\cite{khan2012future, lai_enabling_2018}. The demand for alternatives is clear given the growing number of graph learning applications in resource-constrained scenarios. Examples range from IoT malware detection~\cite{abusnaina2019adversarial} to air pollution monitoring sensor networks \cite{ferrer2021graph}.

In a search for learning methods capable of handling scenarios with limited compute resources, Hyperdimensional computing (HDC) has emerged as an efficient alternative to deep learning~\cite{kanerva2009hyperdimensional}. HDC represents information in a high-dimensional space using \textit{hypervectors}. Each hypervector stores data holographically, that is, each dimension contains the same amount of information. This representation provides inherent robustness to faulty components~\cite{rahimi2016robust}. In addition, the learning algorithms in HDC are based on well-defined operations between hypervectors, which are typically dimension-independent, providing an opportunity for massive parallelism. Unlike the arithmetic-based deep network solutions, HDC's natural expression of massive logic-level parallelism makes it particularly well suited for FPGA, GPU and ASIC mappings. For example, Schmuck et.al.~\cite{schmuck2019hardware} show how the use of associative memories makes it possible to reduce the time of each classification event to the extreme of a single clock cycle. 

The characteristics of HDC have already shown their merit on several problems (see Section~\ref{sec:related}). However, despite the previously stated importance of graph learning applications, to the best of our knowledge, HDC algorithms have never been applied to such tasks. It is based on this motivation that we propose GraphHD, the first of its kind baseline approach for graph learning with HDC. GraphHD focuses on providing an efficient, robust and scalable alternative to current state-of-the-art graph learning algorithms. 

We submit GraphHD to extensive testing on real-world graph classification problems from six publicly available datasets. We compare GraphHD to state-of-the-art graph kernels and GNNs. The comparative results indicate that the method for graph learning based on HDC achieves a comparable accuracy, while being inherently more robust to noise and achieving much higher efficiency compared to GNNs. In addition, experiments are carried out to assess the scalability of the methods in relation to the size of the graphs. GraphHD behaves better than the other methods, achieving training times that are 6.2$\times$ faster than GNNs and 15.0$\times$ faster than kernel methods on the largest graphs. 
\section{Related Work}
\label{sec:related}

Historically, HDC emerged from the cognitive modeling work by Kanerva~\cite{kanerva2009hyperdimensional} and was inspired by characteristics of human memory. Like our memory, the operations on hypervectors were shown to be able to capture information association and bundling. The three fundamental operations—addition (bundling), multiplication (binding), and permutation—were used to create problem-specific encoding algorithms. In early work, the algorithms focused on classifying time series and text sequences~\cite{kleyko2014brain, recchia2015encoding, rahimi2016hyperdimensional, najafabadi2016hyperdimensional, rahimi2016robust, imani2017voicehd}. Recent work has expanded the application opportunities by introducing encoding algorithms for images~\cite{manabat2019performance, imani2019binary} and DNA sequence matching~\cite{kim2020geniehd}.

In addition to the expanding scope of applications and training methods, there has been research into HDC's computational efficiency and robustness. The computational efficiency is a result of the small set of well-defined operations on high-dimensional vectors which allows for massive parallelization~\cite{rahimi2017high, li2016hyperdimensional}. Hardware-level optimizations such as in-memory processing promise to further increase the computational efficiency of HDC~\cite{imani2017ultra}. Schmuck et. al.~\cite{schmuck2019hardware} propose a number of hardware techniques for optimizing HDC operations on an FPGA while simultaneously improving throughput and circuit area. Among other results, they demonstrate that each HDC inference operation can be performed in a single clock cycle. The high-dimensional vectors with random independent and identically distributed (i.i.d.) components make hypervectors a robust way of information representation~\cite{kanerva2009hyperdimensional, wu2018brain}. The opportunity for massive parallel computation in combination with the robust nature of HDC makes it an ideal learning framework for resource constrained environments such as IoT and embedded systems~\cite{benatti2019online}.

With the growing number of applications for machine learning with graphs, several approaches have been proposed in recent years. Popular ones include those based on \textit{graph kernels} and \textit{graph neural networks}. Examples of graph kernel approaches are those based on spectral properties~\cite{kondor2016advances}, random walks~\cite{kang2012fast} and matching of node embeddings~\cite{nikolentzos2017matching}. Some very prominent graph kernels are based on the well-known heuristic for graph isomorphism, i.e., the Weisfeiler-Leman (WL) algorithm~\cite{shervashidze2011weisfeiler,kriege2016valid}. For a detailed and recent review of graph kernels, see Kriege et al.~\cite{kriege2020survey}.

More recently, numerous attempts to adapt neural networks to deal with graphs have come to be known as graph neural networks (GNNs)~\cite{wu2020comprehensive}. The initial concept is due to Gori et al. (2008) \cite{gori2005new}, further elaborated by Scarselli et al. \cite{scarselli2008graph}.
Despite the recent and extensive exploration of GNNs, classical graph kernels are still very competitive in terms of accuracy~\cite{morris2020tudataset} and especially in efficiency (as indicated by our results in Section~\ref{sec:experiments}). Xu et al. \cite{xu2018how} show that GNNs are \textit{at most} as powerful as the WL test in distinguishing graph structures.

While these existing approaches have a well established track record in the field of graph similarity analysis, our HDC approach is a novel attempt at solving graph learning tasks. 
\section{Hyperdimensional Computing}
\label{sec:HDC}
Hyperdimensional computing seeks to emulate brain circuits in a more robust and efficient way than neural networks by representing information as points in a high-dimensional space, called \textit{hypervectors}. Hypervectors are typically binary or bipolar vectors with ten thousand dimensions. Representation and transformations of data in HDC are performed over fixed dimension hypervectors. The cognitive tasks are carried out based on the similarity between those representations.

\begin{figure}[ht]
 \centering
 \includegraphics[width=.8\columnwidth]{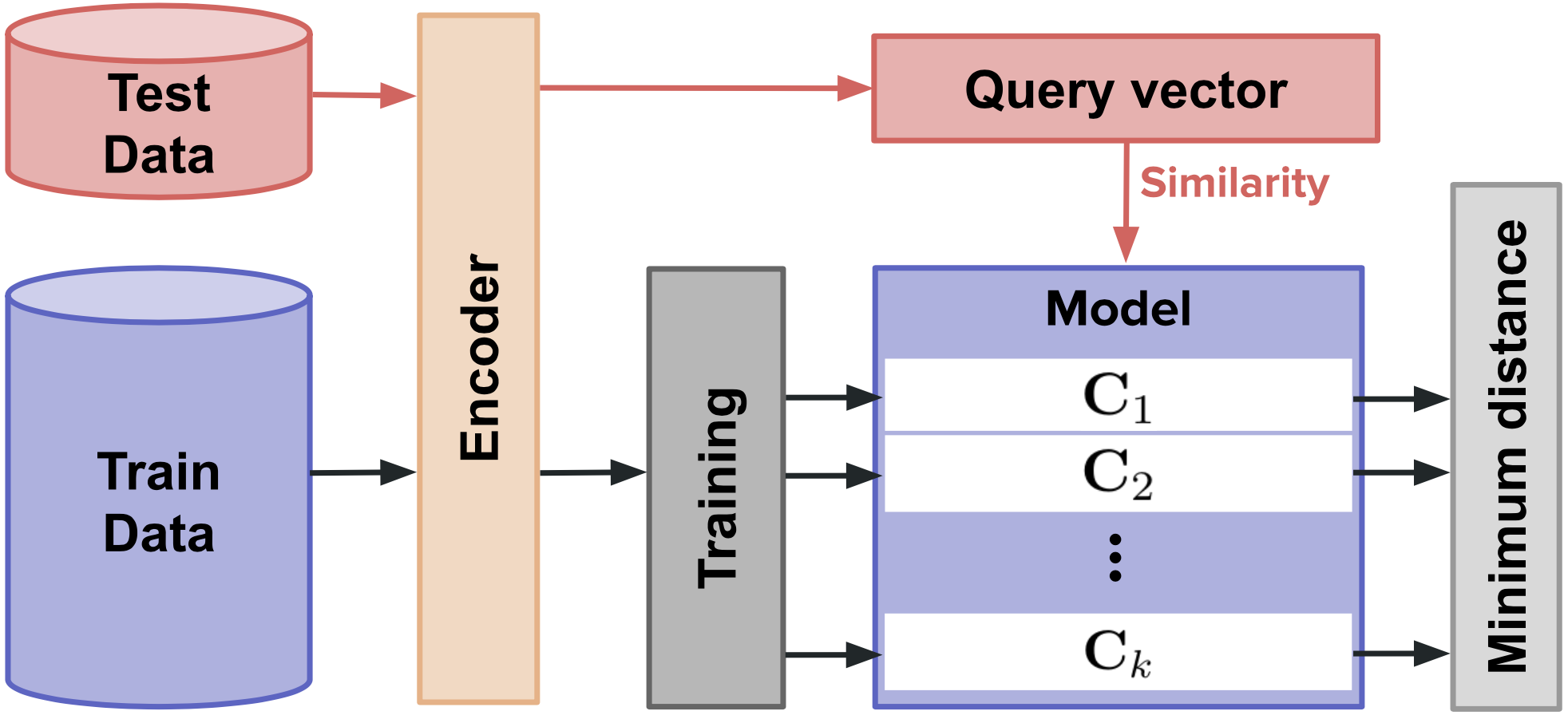}
  \caption{\label{fig:HDC} Hyperdimensional computing classification overview}
\end{figure}

HDC models can often be separated into three stages: encoding, training, and inference. An overview of HDC classification is shown in Figure~\ref{fig:HDC}. The encoding stage is application specific and serves to transform the input data into hypervectors. During the training the hypervectors are aggregated to learn a model. Finally, inferences into the model can be made using the generated class representations. The following sections will cover each stage in more detail.

\subsection{Encoding}
\label{sec:encoding}
The mapping of data to the high-dimensional space is the first step in HDC and this process corresponds to \textit{encoding}. The process is governed by a function $\mathrm{Enc}: {\mathbb I}\to{\mathbb H}^{d}$, that maps input arguments (e.g. graphs, text or images) in $\mathbb I$ to the $d$-dimensional space ${\mathbb H}^d$. Encoding is the HDC counterpart to the feature extraction process in classical learning methods. Thus, the main intuitive principle that governs the encoding is that inputs that are similar in the original space should be mapped to similar hypervectors.

The process starts by generating a set of basis hypervectors that represent units of information (e.g. feature values and positions). The basis hypervectors remain fixed throughout computation and each data sample is encoded by combining and manipulating them using the addition (bundling), multiplication (binding) and permutation operations. 

An example of a commonly used technique is the \textit{record-based encoding}. This encoding binds key and value hypervectors and bundles them to create the hypervector for each data point. The key hypervector can for example correspond to the position in an image or the identifier of an attribute. 

The following equation shows a general example for record-based encoding. The encoding generates the hypervector $\mathbf{H}$ from the randomly generated key hypervectors $\mathbf{K}_i$ which are bound to their value $\bar{\mathbf{V}_i}$ which is one of the predefined value hypervectors in $\mathbf{V}$. The square brackets $[\ldots]$ denote normalization, commonly element-wise majority voting, $\times$ and $+$ respectively represent binding and bundling.
\begin{gather*}
\mathbf{H} = \left[\mathbf{K}_1 \times \bar{\mathbf{V}}_1 + \mathbf{K}_2 \times \bar{\mathbf{V}}_2 + \ldots + \mathbf{K}_N \times \bar{\mathbf{V}}_N\right],\\
\bar{\mathbf{V}}_i \in \{\mathbf{V}_1, \mathbf{V}_2, \ldots, \mathbf{V}_m \}, \mathrm{where} ~1 \leq i \leq N
\end{gather*}


Categorical feature values are mapped to the high-dimensional  space using randomly generated hypervectors, making the hypervectors for each category nearly orthogonal to the others. In the following statement $\delta$ is a given similarity metric between a pair of hypervectors such as the inverse Hamming distance or the cosine similarity.
\begin{gather*}
    \forall i, j:~\delta\left(\mathbf{V}_i, \mathbf{V}_j\right) \simeq 0,~\mathrm{where}~i\neq j
\end{gather*}


\subsection{Training}
\label{sec:training}
The standard HDC training process creates $k$ hypervectors, one for each class. Therefore, a trained model is simply denoted by ${\cal M}=\{\mathbf{C}_1, \mathbf{C}_2, \ldots, \mathbf{C}_k\}$, a set of class-vectors where $\mathbf{C}_i$ contains all the information used to identify the $i$-th class. Each $\mathbf{C}_i$ is calculated as the vector with the smallest average distance to the the hypervectors obtained by encoding the training samples of class $i$:
\begin{align*}
    \mathbf{C}_i = \sum_{j :\ell(x_j )=i} \mathrm{Enc}(x_j )
\end{align*}
where each $x_j \in {\mathbb I}$ and $\ell(x_j ) \in \{1,\dots,k\}$ are a training sample and its respective class. The $\sum$ symbol denotes the element-wise majority voting of hypervectors, named \textit{bundling} (or addition) in HDC.

\subsection{Inference}
\label{sec:inference}
The first step to classify a test sample $y \in {\mathbb I}$ in HDC is to encode $y$. The resulting hypervector $\mathrm{Enc}(y)$ is referred to as query-hypervector. It is important to emphasize that the function $\mathrm{Enc}$ is exactly the same as the one used to encode samples in the training process. After that, the predicted class for $y$ is obtained by checking which class-vector in $\cal M$ is most similar to $\mathrm{Enc}(y)$. In mathematical form:
\begin{align*}
    \mathrm{pred}(y) = \underset{i\in\{1,\dots,k\}}{\arg\max} \;\delta\left(\mathrm{Enc}(y), \mathbf{C}_i \right)
\end{align*}
where $\mathrm{pred}(y)$ is the predicted class for $y$.
\section{GraphHD}
\label{sec:graphHD}

\subsection{Notations and problem formulation}
Let $G=(V,E)$ denote a graph with vertex set $V$ and edge set $E$ with $n=|V|$, $m=|E|$. The class of a specific graph $G$ is denoted by $\ell(G)$. The graph classification problem is defined as follows: given a set of graphs ${\cal G} = \{G_1 , G_2 , \dots ,G_N\}$ and a training subset ${\cal G}_L\subset{\cal G}$ for which the labels are known, create a model capable of predicting the unknown labels for the graphs in ${\cal G}\setminus{\cal G}_L$. 

An important thing to mention is that, since GraphHD was thought of as a baseline method for graph learning with HDC, we assume that graphs do not have any other information such as labels or attributes. Although some datasets contain this type of information, we decided to keep GraphHD as uniform and generic as possible in the present work. The use of this information in ad-hoc applications can be advantageous and shall be investigated in future work.

\subsection{GraphHD overview}
\begin{figure}[ht]
 \centering
 \includegraphics[width=.8\columnwidth]{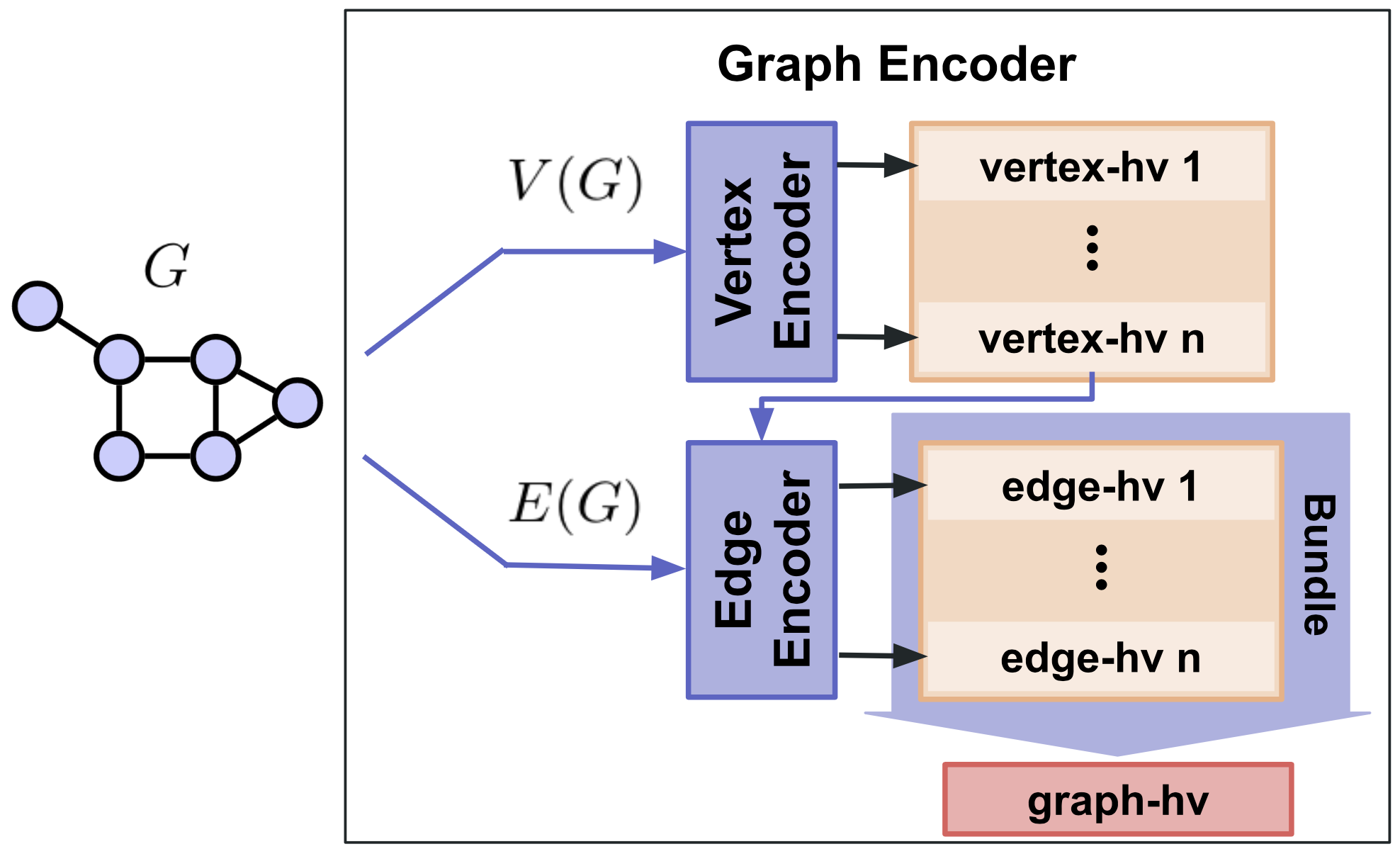}
  \caption{\label{fig:encoding} GraphHD encoding}
\end{figure}

As described in Section~\ref{sec:HDC}, The first and most important question to be addressed when applying HDC to a different domain is: \textit{how to encode the input data?} We want to define a function $\mathrm{Enc}_{G}: {\cal G}\to \mathbb{H}^d$, capable of mapping graphs in the input set to $d$-dimensional hypervectors. Illustrated in Figure~\ref{fig:encoding}, the overall strategy of GraphHD's encoding is to map each element that composes the graph, i.e. its vertices and edges, individually to a hypervector and then combine the information using the bundling operation.

\subsection{GraphHD encoding}
GraphHD first encodes the vertices, those hypervectors are then used to encode the edges. We will first describe the process $\mathrm{Enc}_v :V(G)\to \mathbb{H}^d$ used to encode each element in the set of vertices of a graph $G$, denoted by $V(G)$. In the existing encoding examples, used for non-graph data, the process usually starts by assigning a random hypervector for each possible symbol. For example, one for each letter to encode text.

Based on the existing encoding strategies, one could think of starting to encode graphs by assigning independent random hypervectors to each vertex. However, note that in these other problems, there is a relationship between the symbols that is consistent across different samples of the dataset. For example, the symbol ``A" in a text represents equivalent information in another text, which makes it reasonable to encode both using the same hypervector. However, since we only look at the structure of the graphs, there is no such trivial correspondence between vertices of different graphs.

To address this issue, the vertex encoding process needs to start by extracting an identifier for the vertices based only on the topology of the graph. For this purpose, we propose the use of the \textit{PageRank} centrality metric~\cite{brin1998anatomy}. Initially developed by Google to rank web pages in the web graph, the PageRank algorithm receives a graph as input and returns, for each vertex $v_i \in V$, a value $c(v_i ) \in [0,1]$ that measures its ``importance'' in the graph. The metric is well established and has been widely applied to different problems beyond the web~\cite{gleich2015pagerank}. As rests evident from its initial application, PageRank can be implemented in a very efficient and scalable manner, which matches the purpose of GraphHD.

From this ranking induced by the PageRank centrality of the vertices, it is possible to establish a meaningful connection between vertices in different graphs. Therefore, GraphHD uses the centrality rank of the vertex as its identifier (or symbol). Accordingly, vertices of different graphs, but with the same centrality rank, are encoded to the same random hypervector from the basis set.


After creating the hypervectors for each vertex, GraphHD makes use of these representations to also encode each edge $(v_i , v_j) \in E(G)$. The edge encoding function $\mathrm{Enc}_e$ is defined as follows:
\begin{gather*}
    \mathrm{Enc}_e\left((v_i , v_j ) \right) = \mathrm{Enc}_v (v_i ) \times \mathrm{Enc}_v (v_j )
\end{gather*}
The $\times$ symbol represents the binding operation in HDC, which is the standard operation to represent an association between a pair of hypervectors, similar to the role of an edge in a graph. The result of the binding operation is a third vector, statistically quasi-orthogonal to the operand vectors, which we name \textit{edge-hypervectors}.

\subsection{GraphHD training}

Based on the encoding functions presented, GraphHD training is described in Algorithm~\ref{alg:training}. For each class we generate a set $H_\ell$ of hypervectors. Each hypervector included in $H_\ell$ (line 12 in Algorithm~\ref{alg:training}) is what we call a \textit{graph-hypervector}. For each graph $G$, the corresponding graph-hypervector is created with $\mathrm{bundle}(H_G )$, which bundles all the edge-hypervectors contained in the set $H_G$. Note that vertex encoding is used as an intermediate edge encoding step as defined above.

\begin{algorithm}
    \label{alg:training}
    \SetKwInOut{Input}{Input}
    \SetKwInOut{Output}{Output}

    \underline{GraphHD\_Training} $({\cal G}_L)$\;
    \Input{A training set of graphs ${\cal G}_L$ with their respective labels and a set of random vertex-hypervectors $H_v$.}
    \Output{A trained HDC model ${\cal M}$ consisting of the class vectors $\{\mathbf{C}_1, \ldots, \mathbf{C}_k\}$}
    \For{each class label $i\in\{1, \ldots, k\}$}
    {
        $H_\ell \gets \varnothing$\\
        \For{each graph $G\in{\cal G}_L: \ell(G)=i$}
        {
            $H_G \gets \varnothing$\\
            \For{each edge $e \in E(G)$}
            {$H_G \gets H_G \cup \mathrm{Enc}_e (e)$}
            $H_\ell \gets H_\ell \cup \mathrm{bundle}(H_G)$
        }
        $\mathbf{C}_i \leftarrow \mathrm{bundle}(H_\ell )$
    }
    {
        \Return $\{\mathbf{C}_1, \ldots, \mathbf{C}_k\}$
    }
    \caption{GraphHD training procedure}
\end{algorithm}
\section{Experiments}
\label{sec:experiments}

\begin{figure*}[h]
    \centering
    \includegraphics[width=\linewidth]{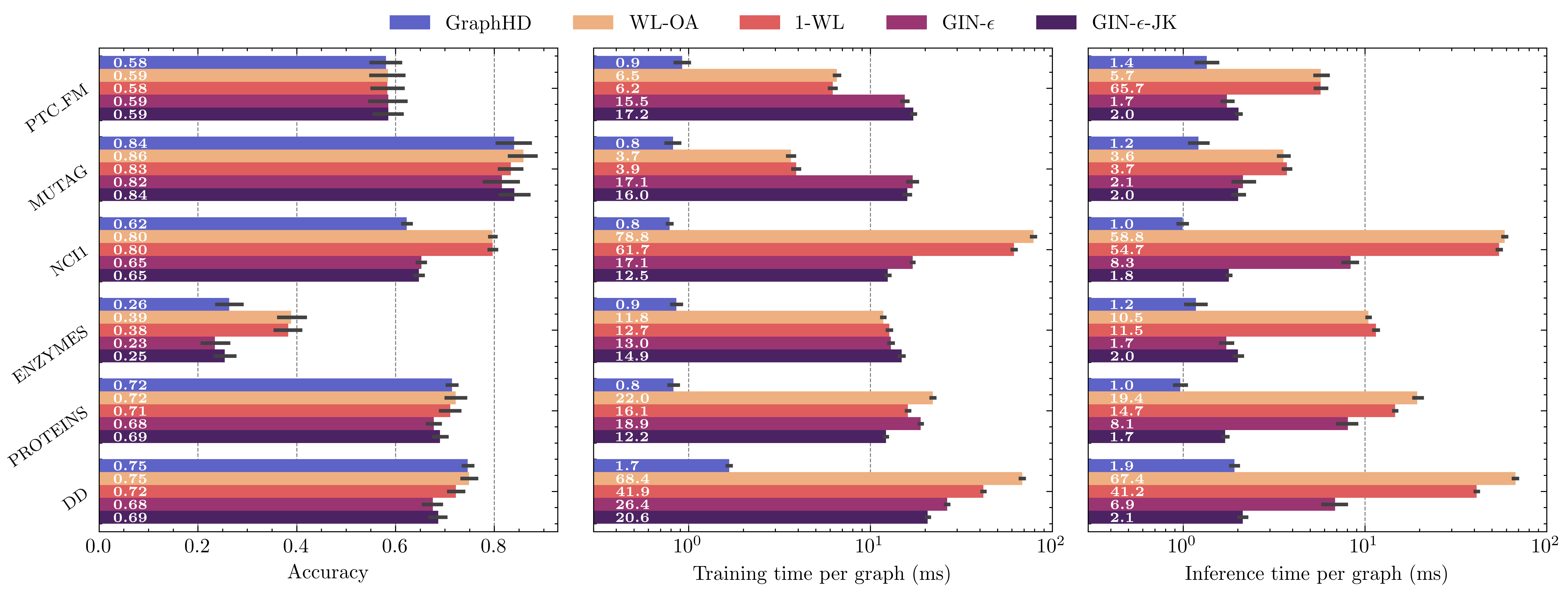}
    \caption{\label{fig:accuracy} \textbf{Left:} Accuracy achieved on six datasets by GraphHD compared to that of the kernel methods, 1-WL and WL-OA, and the graph neural networks, GIN-$\epsilon$ and GIN-$\epsilon$-JK. \textbf{Middle:} Training time in seconds of the five learning methods for each of the six datasets. Note that the y-axis is in logarithmic scale. \textbf{Right:} Average inference time for a graph in each of the six datasets compared across the five learning methods. Note that the y-axis is in logarithmic scale.}
\end{figure*}

To evaluate GraphHD, two groups of experiments were conducted. First, we adopt six graph classification datasets to evaluate the accuracy, training times, and inference times. All of these benchmarks are part of TUDataset, a collection of datasets and standardized evaluation procedures widely used in the empirical evaluation of graph classification methods~\cite{morris2020tudataset}. The performance of GraphHD is compared to two kernel methods and two graph neural networks. Secondly, the scaling profile of GraphHD is empirically assessed and presented in comparison to a GNN and a kernel method. 

All the experiments are performed on the same machine with 20 Intel Xeon Silver 4114 CPUs at 2.20GHz with 93~GB of RAM and 4 Nvidia TITAN Xp GPUs with 3840 cores and 12~GB running CentOS Linux release 7.9.2009.

GraphHD uses 10,000-dimensional bipolar hypervectors in all the experiments. We fix the number of PageRank iterations to 10 for all experiments because the accuracy of GraphHD has then plateaued. The PageRank batch size is 256.

\subsection{Accuracy and Training time}
\label{sec:acc-train-time}
We compare the accuracy and the training and inference times of GraphHD on six datasets from the TUDataset~\cite{morris2020tudataset} collection against four methods. We use 10-fold cross validation because the datasets contain relatively few graphs. We report training and inference time per graph to normalize over varying dataset lengths. The wall-time for one fold of training is considered the training time. The inference time is set to be the testing wall-time of one fold. Measurements are averaged over 3 repetitions of 10-fold cross validation.

GraphHD is expected to significantly outperform existing methods on training time. The experiment gives insight into exactly how much faster and how much accuracy is traded-off for the increase in training speed. Since GraphHD only takes the structure of the graph into account we restrict the GNNs and kernel methods from utilizing the vertex and edge labels.

\subsubsection{Datasets}
\label{sec:datasets}


An overview of the statistics of the datasets used is shown in Table~\ref{tab:dataset-overview}. ENZIMES~\cite{borgwardt2005protein} is a dataset of protein structures, and the task is to assign each enzyme to one of 6 Enzyme Commission (EC) top-level classes. MUTAG~\cite{debnath1991structure} has graphs representing mutagenic aromatic and heteroaromatic nitro compounds with 7 labels. NCI1~\cite{wale2006comparison} is a set of data from the National Cancer Institute (NCI) and the task is to classify chemical compounds according to their ability to inhibit cancerous cell multiplication. In the graph datasets PROTEINS~\cite{dobson2003distinguishing} and DD~\cite{dobson2003distinguishing}, the task is to classify whether or not the represented proteins are non-enzymes. Finally, PTC\_FM~\cite{helma2001predictive} is a dataset from the Predictive Toxicology Challenge containing a collection of chemical compounds represented as graphs which report the carcinogenicity for rats. 

The selected datasets, with the exception of DD, contain very small graphs, with an average of less than 40 vertices. The graphs in the selected datasets are also very sparse, the average fraction of connected vertices is 0.05. The dataset containing the largest graphs, DD, should give an indication on real data of how well the learning methods scale.

\begin{table}[h]
\centering
\caption{\label{tab:dataset-overview} Statistics of graph datasets.}
\begin{tabular}{lllll}
\hline
\textbf{Dataset}     & \textbf{Graphs} & \textbf{Classes} & \textbf{Avg. vertices} & \textbf{Avg. edges} \\ 
\hline
DD          & 1178	 & 2	   & 284.32	    & 715.66     \\
ENZYMES     & 600    & 6       & 32.63      & 62.14      \\
MUTAG       & 188    & 2       & 17.93      & 19.79      \\
NCI1        & 4110   & 2       & 29.87      & 32.3       \\
PROTEINS    & 1113   & 2       & 39.06      & 72.82      \\
PTC\_FM     & 349    & 2       & 14.11      & 14.48      \\
\hline
\end{tabular}
\end{table}

\subsubsection{Baselines}
\label{sec:baselines}

As baseline methods for comparison, two state-of-the-art GNNs and two kernel-based methods were used. The methods are the most recently published that are available for standardized evaluation in the TUDataset repository. The two selected GNN methods are GIN-$\epsilon$~\cite{xu2018how} and GIN-$\epsilon$-JK~\cite{xu2018representation}. For both GNN methods the network architecture was fixed for all experiments at 1 layer with 32 units. We found that this is the smallest network size that matches or exceeds the accuracy of GraphHD on all datasets. We use the Adam optimizer with a learning rate scheduler starting at 0.01 with a patience parameter of 5 which decays with 0.5 till a minimum of $10^{-6}$, and the batch size is 128. For the kernel methods baseline \textit{Weisfeiler-Lehman Subtree}~(1-WL)~\cite{shervashidze2011weisfeiler} and \textit{Weisfeiler-Lehman Optimal Assignment}~(WL-OA)~\cite{kriege2016valid} were used. As part of the training process the C-parameter of the kernels are selected from $\left\{10^{-3}, 10^{-2}, \ldots, 10^2, 10^3\right\}$ and the number of iterations from $\left\{0, \ldots, 5\right\}$. The training hyper-parameters, except for the fixed size GNN architecture, were taken from the reference baseline experiments.

\subsection{Scalability}
To confirm the superior computational efficiency of HDC, the scalability experiment looks at how training time relates to the size of the graphs in the dataset. We create synthetic datasets with 2 classes evenly split over 100 graphs with varying numbers of vertices using the Erdős–Rényi random graph model~\cite{gilbert1959random}. The edge probability is set to 0.05, which is similar to the average connections in the real-world datasets as derived from the dataset statistics in Table~\ref{tab:dataset-overview}. GraphHD is compared against one GNN and one kernel method, GIN-$\epsilon$ and WL-OA, respectively. The methods use the same hyper-parameters as described in Section \ref{sec:baselines}.

\section{Results and discussion}
\label{sec:results}

\begin{figure}[htb]
 \centering
 \includegraphics[width=0.7\linewidth]{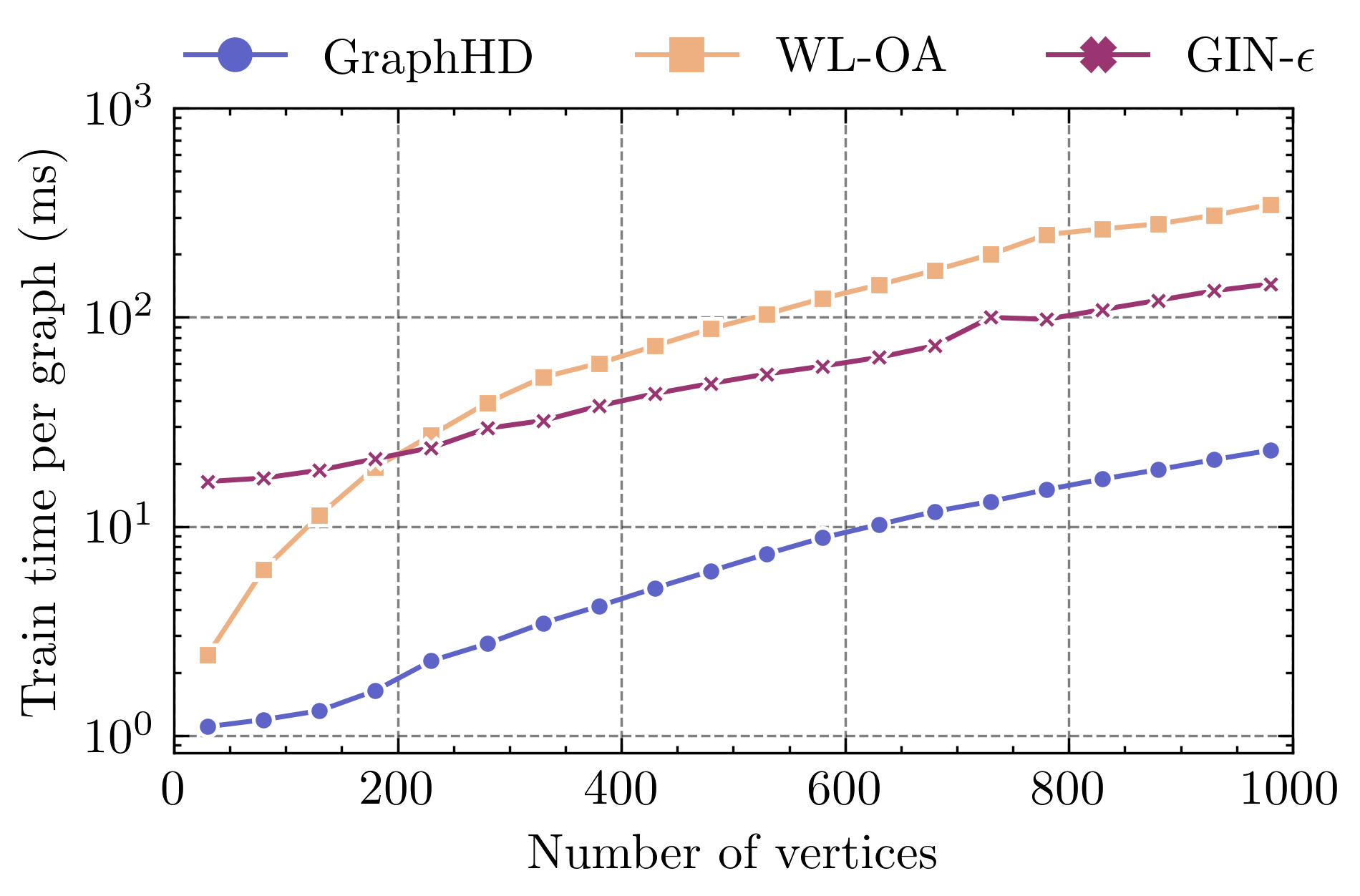}
 \caption{\label{fig:scalability} Scaling profile of GraphHD compared to the graph neural network GIN-$\epsilon$ and the kernel method WL-OA. The y-axis is in logarithmic scale.}
\end{figure}

The accuracy results, shown on the left in Figure~\ref{fig:accuracy}, show that GraphHD has comparable accuracy to the baseline methods, except for NCI1 and ENZYMES where the kernel methods respectively do 18\% and 12\% better than both GraphHD and the GNN methods.

The training time results, shown in the middle of Figure~\ref{fig:accuracy}, confirm the notion that HDC is more computationally efficient than the other learning methods. GraphHD trains significantly faster than both the kernel and GNN methods on every dataset. 
On the DD dataset, which contains the largest graphs, GraphHD trains 12.1$\times$ faster than the GNNs and 24.6$\times$ faster than the fastest kernel method. Moreover, on the largest dataset, NCI1, GraphHD trains 77.1$\times$ faster than the kernel methods. Confirming that with respect to the dataset size the kernel methods have inferior scaling.

The inference time of GraphHD, shown on the right in Figure~\ref{fig:accuracy}, is also faster than the other methods for every dataset. On the DD dataset the fastest GNN is 10.5\% slower and the kernel methods are 21.7$\times$ slower.  

The scaling profile of GraphHD, as shown in Figure \ref{fig:scalability}, is up to an order of magnitude lower than that of the baseline methods as the graphs' sizes increase. At the largest measured graphs with 980 vertices, GraphHD is 6.2$\times$ faster than GIN-$\epsilon$ and 15.0$\times$ faster than WL-OA. The faster training and inference times allow for more graph learning applications to become feasible. These findings are consistent with those from the training times on the real-world datasets from the accuracy and training time experiment described in Section~\ref{sec:acc-train-time}. It is worth to remind the potential of HDC dedicated hardware to further improve the training and inference times of GraphHD as discussed in Section~\ref{sec:related}.

\section{Future Work}
\label{sec:future_work}

As a first HDC approach for graph classification, a goal of GraphHD is to demonstrate the effectiveness of the idea and thereby open and motivate new research directions and advance the problem frontier. Some of the main directions that may be further investigated include: 1) To what extent is it possible to sacrifice the efficiency of GraphHD to match and possibly surpass the accuracy of the other methods? This could be done by extending GraphHD with techniques already known in HDC such as retraining and multiple class-vectors per class;
2) As previously mentioned, in some specific applications, the datasets contain information other than the graph structure, such as labels and attributes for both vertices and edges. It would be interesting to study methods that incorporate these types of additional information into GraphHD.
\section{Conclusion}
\label{sec:conclusion}
We show that GraphHD achieves comparable accuracy while proving to be significantly more efficient. Remarkably when the graphs increase in size, the scaling profile of GraphHD is much more favorable, opening up possibilities for graph classification on large graphs that were previously not computationally feasible. We introduced a baseline graph encoding algorithm that makes it possible to use HDC for graph learning applications. The results of GraphHD are promising and indicate the importance of continuing to investigate HDC algorithms for graph learning as a light-weight, robust and scalable alternative to deep learning, especially in resource constrained applications such as embedded devices and IoT.
\bibliographystyle{IEEEtran}
{\linespread{0.9}\selectfont\bibliography{references_compact}}

\end{document}